\documentclass
[%
	10pt,  % [9pt, >10pt, 11pt, 12pt]
	draftcls,  % [draft, draftcls, draftclsnofoot, >final] cls=yes figures, nofoot=no draft footnote
	peerreviewca,  % [conference, >journal, technote, peerreview, peerreviewca] ca=for conference mode
]%
{IEEEtran}%

\IEEEoverridecommandlockouts%
\usepackage{url}

\usepackage{cite}		% some citation fixes, e.g., sorting citations
\usepackage{graphicx}	% figure commands
\usepackage[table]{xcolor}		% color commands
\usepackage{			% math typesetting commands
	amsmath,%
	amssymb,%
	amsfonts,%
}%
\usepackage[noend]{algpseudocode}	% Algorithm Boxes, see acronyms.tex for commands
\usepackage[%
	nohyperlinks,		% disable hyperlinks
	nolist,				% dont print acronym list
]{acronym}

\usepackage{siunitx}	% Units with correct spacing. \SI{12}{\dB}, \SI{4}{\mega\hertz}, \ang{90}, \num{1e-3}, \num{10000}
\usepackage{xspace}		% Intelligent spacing after text macros
\usepackage{tikz}
\usetikzlibrary{%
	positioning,				% relative positioning
	shapes.geometric,			% circle, ellipse etc
	decorations.pathreplacing,	% curly braces etc
	shadows,					% drop shadows
	patterns,					% patterns
}%

\DeclareGraphicsExtensions{.pdf,.jpeg,.jpg,.png}
\input{99_custommacros.sty}
\newcommand{\timesteptxt}{time step~\( \timeindex \)\xspace}
\newcommand{\jobtxt}{job~\( \jobindex \)\xspace}
\newcommand{\usertxt}{user~\( \userindex \)\xspace}

\newcommand{\probability}								{p}
\newcommand{\probnewjob}{\probability_{\text{job}}}
\newcommand{\probpriorityjob}{\probability_{\text{prio}}}

\newcommand{\userindex}									{n}
\newcommand{\totalusers}								{N}

\newcommand{\channelstateuser}[1][\userindex]{\internalchannelcoefficient_{#1, \timeindex}}
\newcommand{\variancechannel}{\variance_{\text{R}}}

\newcommand{\resourceblock}{u}
\newcommand{\totalresourceblocks}{U}
\newcommand{\resourceblocksrequesteduser}[1][\userindex]{\resourceblock_{#1, \text{req}, \timeindex}}
\newcommand{\resourceblocksrequestedjob}[1][\userindex]{\resourceblock_{\jobindex, \text{req}, \timeindex}}
\newcommand{\resourcesscheduleduser}{\resourceblock_{\text{sch},\userindex, \timeindex}}
\newcommand{\resourceblocksrequesteduserinitial}[1][\userindex]{\resourceblock_{#1, \text{init}}}
\newcommand{\resourceblocksizesmax}{\resourceblock_{\text{max}}}

\newcommand{\jobindex}									{j}
\newcommand{\jobset}									{\mathbb{J}}
\newcommand{\jobrequestedset}[1][\timeindex]{\jobset_{\text{req}, #1}}
\newcommand{\jobrequesteduserset}[1][\userindex]{\jobset_{\text{req}, #1, \timeindex}}
\newcommand{\jobstimedoutset}{\jobset_{\text{TO}, \timeindex}}
\newcommand{\jobstimedoutpriorityset}{\jobset_{\text{TO,prio}, \timeindex}}

\newcommand{\delay}										{d}
\newcommand{\jobdelay}{\delay_{\jobindex, \timeindex}}
\newcommand{\jobdelaymax}{\delay_{\text{max}}}

\newcommand{\actionvec}{\mathbf{\internalaction}_{\timeindex}}
\newcommand{\actionvecnoisy}{\tilde{\mathbf{\internalaction}}_{\timeindex}}
\newcommand{\explorationvec}{\bar{\mathbf{\internalaction}}_{\timeindex}}
\newcommand{\explorationvecsca}{\bar{\internalaction}_{\userindex, \timeindex}}
\newcommand{\statevec}[1][\timeindex]{\mathbf{\internalstate}_{#1}}
\newcommand{\statescauser}[1]{\internalstate_{\userindex, #1, \timeindex}}
\newcommand{\percentageuser}[1][\userindex]{a_{#1}}

\newcommand{\explorationmomentum}{\internalexplorationparam_{\text{expl}}}

\newcommand{\parameterindex}							{i}
\newcommand{\totalparameters}							{I}
\newcommand{\loss}										{L}
\newcommand{\learningrate}								{\lambda}
\newcommand{\actornetworkvec}{\mathbf{\internalactor}}
\newcommand{\parameteractor}{\internalparameter_{\internalactor, \timeindex}}
\newcommand{\parametersactor}{\mathbf{\internalparameter}_{\internalactor, \timeindex}}
\newcommand{\parameteractoranchor}{\internalparameter_{\internalactor, \parameterindex, \timeindex}^{\text{EWC}}}
\newcommand{\criticnetworksca}{\internalcritic}
\newcommand{\parameterscritic}{\mathbf{\internalparameter}_{\internalcritic, \timeindex}}
\newcommand{\lossactor}{\mathcal{\loss}_{\actornetworkvec}}
\newcommand{\losscritic}{\mathcal{\loss}_{\criticnetworksca}}
\newcommand{\lossewc}{\mathcal{\loss}_{\text{EWC}}}
\newcommand{\fisherinformation}{F}
\newcommand{\fisherinformationparameter}{\fisherinformation_{\parameterindex}^{\text{EWC}}}

\newcommand{\totalepisodes}								{E}
\newcommand{\totaltimesteps}							{T}
\newcommand{\reward}{\internalreward_{\timeindex}}
\newcommand{\rewardsumrate}{\internalreward_{\text{C}, \timeindex}}
\newcommand{\rewardtimeout}{\internalreward_{\text{TO}, \timeindex}}
\newcommand{\rewardtimeoutprio}{\internalreward_{\text{TO,prio}, \timeindex}}
\newcommand{\rewardestimate}{\hat{\internalreward}_{\timeindex}}
\newcommand{\weightsumrate}{\internalweight_{\text{C}}}
\newcommand{\weighttimeout}{\internalweight_{\text{TO}}}
\newcommand{\weighttimeoutprio}{\internalweight_{\text{TO,prio}}}
\newcommand{\weightanchor}								{\eta}

\newcommand{\gemtotalnumsamples}						{K}
\newcommand{\gradient}									{g}
\newcommand{\gemgradientold}{\mathbf{\gradient}_{\text{prio}}}
\newcommand{\gemgradientcurrent}{\mathbf{\gradient}_{\text{curr}}}
\newcommand{\gemgradientnew}{\tilde{\mathbf{\gradient}}_{\text{curr}}}
\newcommand{\gemoptimizationvariable}					{v}
\newcommand{\gemoptimizationvariableoptimal}{\gemoptimizationvariable^{\text{opt}}}

\begin{document}

\title
{%
%	Increasing Robustness Against Black Swan Events in Deep Reinforcement Learned Resource Allocation: A Multi-Task Approach%
	A Multi-Task Approach to Robust Deep Reinforcement Learning for Resource Allocation
	\thanks{This work was partly funded by the German Ministry of Education and Research (BMBF) under grant 16KIS1028 (MOMENTUM) and 16KISK016 (Open6GHub) and the European Space Agency (ESA) under number 4000139559/22/UK/AL.}%
	\thanks{The contents of work were presented on the WSA/SCC2023.}
}%
\author{%
	\IEEEauthorblockN{%
	Steffen~Gracla%
	%\,\orcidlink{0000-0003-3315-9280}%
	,
	Carsten~Bockelmann%
	%\,\orcidlink{0000-0002-8501-7324}%
	,
	and Armin~Dekorsy%
	%\,\orcidlink{0000-0002-5790-1470}%
	}%
	\IEEEauthorblockA{%
		Dept. of Communications Engineering, University of Bremen, Bremen, Germany\\
		{Email: \{gracla, bockelmann, dekorsy\}@ant.uni-bremen.de}
	}%
}%

%
%\author
%{%
%	\IEEEauthorblockN{1\textsuperscript{st} Given Name Surname}
%	\IEEEauthorblockA{\textit{dept. name of organization (of Aff.)} \\
%	\textit{name of organization (of Aff.)}\\
%	City, Country \\
%	email address or ORCID}
%	\and
%	\IEEEauthorblockN{2\textsuperscript{nd} Given Name Surname}
%	\IEEEauthorblockA{\textit{dept. name of organization (of Aff.)} \\
%	\textit{name of organization (of Aff.)}\\
%	City, Country \\
%	email address or ORCID}
%	\and
%	\IEEEauthorblockN{3\textsuperscript{rd} Given Name Surname}
%	\IEEEauthorblockA{\textit{dept. name of organization (of Aff.)} \\
%	\textit{name of organization (of Aff.)}\\
%	City, Country \\
%	email address or ORCID}
%	\and
%	\IEEEauthorblockN{4\textsuperscript{th} Given Name Surname}
%	\IEEEauthorblockA{\textit{dept. name of organization (of Aff.)} \\
%	\textit{name of organization (of Aff.)}\\
%	City, Country \\
%	email address or ORCID}
%	\and
%	\IEEEauthorblockN{5\textsuperscript{th} Given Name Surname}
%	\IEEEauthorblockA{\textit{dept. name of organization (of Aff.)} \\
%	\textit{name of organization (of Aff.)}\\
%	City, Country \\
%	email address or ORCID}
%	\and
%	\IEEEauthorblockN{6\textsuperscript{th} Given Name Surname}
%	\IEEEauthorblockA{\textit{dept. name of organization (of Aff.)} \\
%	\textit{name of organization (of Aff.)}\\
%	City, Country \\
%	email address or ORCID}
%}%
%
\maketitle%
%

%%%%%%%%%%%%%%%%%%%%%%%%%%%%%%%%%%%%%%%%%%%%%%%%%%%%%%%%%%
% REMEMBER STORYTELLING
%%%%%%%%%%%%%%%%%%%%%%%%%%%%%%%%%%%%%%%%%%%%%%%%%%%%%%%%%%

\begin{abstract}
	With increasing complexity of modern communication systems, \ac{ml} algorithms have become a focal point of research.
	However, performance demands have tightened in parallel to complexity.
	For some of the key applications targeted by future wireless, such as the medical field, strict and reliable performance guarantees are essential, but vanilla \ac{ml} methods have been shown to struggle with these types of requirements.
	Therefore, the question is raised whether these methods can be extended to better deal with the demands imposed by such applications.
	In this paper, we look at a combinatorial \ac{ra} challenge with rare, significant events which must be handled properly. We propose to treat this as a multi-task learning problem, select two methods from this domain, \ac{ewc} and \ac{gem}, and integrate them into a vanilla actor-critic scheduler.
	We compare their performance in dealing with Black Swan Events with the state-of-the-art of augmenting the training data distribution and report that the multi-task approach proves highly effective.
\end{abstract}


\acresetall  % Resets the "first use" counter of acronym package
%

%%%%%%%%%%%%%%%%%%%%%%%%%%%%%%%%%%%%%%%%%%%%%%%%%%%%%%%%%%
% REMEMBER STORYTELLING
%%%%%%%%%%%%%%%%%%%%%%%%%%%%%%%%%%%%%%%%%%%%%%%%%%%%%%%%%%

\section{Introduction}
\label{sec:introduction}

As the journey towards the 6th generation 3GPP mobile communication standard continues, many of the most noteworthy targeted use cases are becoming highly specific, with requirements more extreme and heterogeneous~\cite{david_6g_2018, ngmn_6g_2021}.
While the average user may be satisfied with the service provided by 5G NR, as it is currently being implemented by telecommunications service providers world wide, application fields such as medical communications have demands in, \eg bit rate, latency, and reliability that are edge cases even in 5G NR~\cite{3gpp_service_2021, cisotto2020requirements}.
Among the many challenges faced in implementing efficient communication networking, the task of \ac{ra} is among those that may limit performance the most.
A capable resource scheduler must consider and balance available information, demands, and potential constraints in order to find an optimal allocation solution.
For applications with heterogeneous service demands, the combinatorial problems found in resource allocation can quickly become burdensome to solve optimally in real time~\cite{xu_survey_2021}.
For this reason, \ac{ra} has attracted significant interest in applying \ac{ml}~\cite{zhang_deep_2019}.

\ac{ml}, and, in particular, \ac{dl}, have recently demonstrated strong performance in data driven function approximation in multiple domains, remarkably so in language and image processing, which are traditionally considered highly complex to solve well.
Their strength lies in the ability to infer an approximately optimal solution from a given data set without the need to try to model the underlying processes.
Instead, for example, the subdomain of \ac{rl} learns by ``trial and error'', roughly trying to increase the likelihood of decisions that show to produce useful results while avoiding those that do not.
First results in applying \ac{rl} to the task of \ac{ra} in wireless communication have shown promising results, \eg~\cite{roshdi_deep_2021, eisen_learning_2019, kasgari_experienced_2020}.
For especially demanding applications, however, some issues remain with these approximated algorithms.
While a visual glitch may be frustrating in image generation, a missed emergency transmission may prove fatal.
Unfortunately, some of the more potent \ac{ml} algorithms have been shown to struggle learning from data samples that are underrepresented in the overall data set~\cite{fujimoto_addressing_2018}.

A typical approach to helping this issue in communication systems is to artificially increase the relative amount of priority events in the learning data set, by, \eg influencing data generation~\cite{kasgari_experienced_2020} or jointly training on two data sets~\cite{chae_autonomous_2017} with tailored distributions.
Experience from domains like autonomous driving shows that algorithms trained this way, \ie trained on a data distribution that does not match reality, may struggle to reach expected performance when applied to the desired application~\cite{pinto_robust_2017}.
Further, with more complex applications, finding and generating the correct augmented data set that induces the desired behavior in the learned algorithm becomes a taxing and non-intuitive task.
Lastly, these methods have no protection against \emph{catastrophic forgetting}~\cite{kirkpatrick_overcoming_2017}, \ie overwriting desired behavior if learning is continued on new data with a lack of priority events.

We present a procedure to harden \ac{rl} schedulers in the presence of rare events by a multi-task learning approach.
Considering the task of discrete resource scheduling on the MAC layer, the learned scheduler will have to solve three conflicting objectives: 1)~Maximizing Sum-Rate, 2)~minimizing time outs, and especially 3) minimizing time outs on rare events that demand very low latency.
We show that, by default, the scheduler is not able to learn proper management of these rare priority events.
Therefore, we make use of methods from multi-task learning, where a single learned algorithm must sequentially learn multiple different tasks with limited loss in performance on earlier tasks.
We define our first task, that we wish to learn to satisfaction and then prevent unlearning, as the handling of priority messages with high reliability.
Subsequently, we then integrate two methods from multi-task learning with a vanilla deep \ac{rl} resource scheduler which we used in~\cite{gracla_scheduling_2022}.
We show that the two methods,
\begin{enumerate}
	\item \ac{ewc}~\cite{kirkpatrick_overcoming_2017}, presented in our work~\cite{gracla_robust_2022}.
	\ac{ewc} imposes an elastic penalty onto the learning objective, incentivizing learning steps that cause the least change in expected behavior on scheduling priority events by way of Fisher information;
	\item \ac{gem}~\cite{lopez-paz_gradient_2017} proposes to foster positive transfer learning between tasks by only allowing learning steps that do not cause negative impact on a representative set of priority event data samples,
\end{enumerate}
are competitive in performance to the default approach of augmenting the learning data distribution while also compensating the issues mentioned earlier.

In the following, we will start by introducing the \ac{ra} system model and optimization objective.
We then recapitulate the design of a deep \ac{rl} scheduler, and detail the workings of the two multi-task learning methods and their integration with the \ac{rl} scheduler.
We then evaluate and discuss their performance, advantages, and drawbacks.

\emph{Preliminaries \& Notation}: This paper assumes knowledge of stochastic gradient descent optimization and feed-forward \acp{nn}.
Vectors and matrices are denoted in boldface (\( \mathbf{x} \)), sets in blackboard-boldface (\( \mathbb{N} \)).
%

%%%%%%%%%%%%%%%%%%%%%%%%%%%%%%%%%%%%%%%%%%%%%%%%%%%%%%%%%%
% REMEMBER STORYTELLING
%%%%%%%%%%%%%%%%%%%%%%%%%%%%%%%%%%%%%%%%%%%%%%%%%%%%%%%%%%

\section{System Model}
\label{sec:setup}

In this section, we will first introduce the specifics of the discrete frequency allocation problem with significant rare events.
We then formulate a heterogeneous utility function that is to be optimized.

\subsection{Discrete Resource Allocation}
\label{sec:systemmodel}

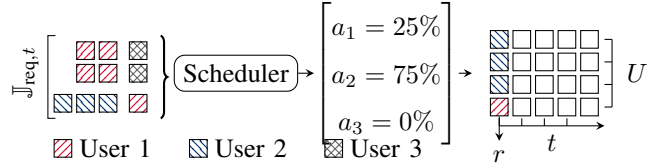
\begin{figure}[!t]
	\centering
	\begin{tikzpicture} [scale=1.0]
\tikzstyle{A0} = [-, >={stealth}, rounded corners]
\tikzstyle{A1} = [->, >={stealth}, rounded corners]
\tikzstyle{A2} = [<->, >={stealth}, rounded corners]
\tikzstyle{selfloop} = [looseness=4]
\tikzstyle{rb} = [draw, minimum width = .1cm, minimum height = .1cm]

\newcommand{\xshift}{0.3}
\newcommand{\yshift}{0.3}

% RESOURCE GRID------------------------------------------------------------------------------
\node (rb11)
	at (+0.0 * \xshift, +1.5*\yshift)
	[rb, pattern=north east lines, pattern color=unired2]
	{};
\node (rb12)
	at (+0.0 * \xshift, +2.5*\yshift)
	[rb, pattern=north west lines, pattern color=uniblue1]
	{};
\node (rb13)
	at (+0.0 * \xshift, +3.5*\yshift)
	[rb, pattern=north west lines, pattern color=uniblue1]
	{};
\node (rb14)
	at (+0.0 * \xshift, +4.5*\yshift)
	[rb, pattern=north west lines, pattern color=uniblue1]
	{};
	
\foreach \columnid in {1.5, 2.5, ..., 4.5}
{
	\foreach \rowid in {1, 2, ..., 4}
	{
		\node (rb\rowid\columnid)
		at (+\rowid * \xshift, +\columnid * \yshift)
		[rb]
		{};
	}
}

\draw []
	(1.5, 1.35)
	-- node [right=.2em] {$\totalresourceblocks$}
	++(0.0, -3 * \yshift)
	;
	
\foreach \x in {0, ..., 3}
	\draw []
		(1.5, 1.35 - \x * \yshift)
		--
		++(-0.1, 0.0);

\draw [A1]
	(0.0, 0.2)
	-- node [below, align=center] {\( \timeindex \)}
	++(1.4, 0.0);

\foreach \x in {0, ..., 3}
	\draw []
	(0.0 + \x * 1.2 / 4, 0.2) -- ++ (0.0, 0.1);
	
\node (scheduler)
	at (-3.5, 0.9)
	[draw, rounded corners]
	{Scheduler};

\node (solution)
	at (-1.5, 0.9)
	[align=left, outer sep=0pt, inner sep=0pt]
	{$
		\begin{bmatrix}
		\percentageuser[1] = 25\%\\
		\percentageuser[2] = 75\%\\
		\percentageuser[3] = 0\%
		\end{bmatrix}
	$};

\node (reward)
	at (0.0, -0.2)
	[]
	{\( \internalreward \)};
	
\draw [A1]
	(rb11)
	to
	(reward);

\draw [A1]
	(scheduler)
	to
	(solution);
	
\draw [A1]
	(solution.east)
	to
	++ (right:.5em);

\draw
	[decorate, decoration={brace}, thick]
	(-4.5, 1.4)
	--
	(-4.5, 0.3);

% JOBS IN REQUESTS-------------------------------------------------------
\foreach \columnid in {0, 1}
{
	\foreach \rowid in {0, 1}
	{
		\node (rb\rowid\columnid)
			at (-5.5 +\rowid * \xshift, 0.9 +\columnid * \yshift)
			[rb, pattern=north east lines, pattern color=unired2]
			{};
	}
}

\foreach \columnid in {0, 1}
{
	\foreach \rowid in {0}
	{
		\node (rb\rowid\columnid)
			at (-4.8 +\rowid * \xshift, 0.9 +\columnid * \yshift)
			[rb, pattern=crosshatch, pattern color=black!60]
			{};
	}
}

\foreach \columnid in {1}
{
	\foreach \rowid in {0}
	{
		\node (rb\rowid\columnid)
			at (-4.8 +\rowid * \xshift, +0.2 +\columnid * \yshift)
			[rb, pattern=north east lines, pattern color=unired2]
			{};
	}
}

\foreach \columnid in {1}
{
	\foreach \rowid in {0, 1, 2}
	{
		\node (rb\rowid\columnid)
			at (-5.8 +\rowid * \xshift, +0.2 +\columnid * \yshift)
			[rb, pattern=north west lines, pattern color=uniblue1]
			{};
	}
}

% USER TEXT--------------------------------------------------------------
\node (user1)
	at (-5.8, -0.1)
	[rb, pattern=north east lines, pattern color=unired2]
	{};
	
\node (user1text)
	at (-5.1, -0.1)
	[]
	{User 1};
	
\node (user2)
	at (-4.0, -0.1)
	[rb, pattern=north west lines, pattern color=uniblue1]
	{};

\node (user2text)
	at (-3.3, -0.1)
	[]
	{User 2};

\node (user3)
	at (-2.2, -0.1)
	[rb, pattern=crosshatch, pattern color=black!60]
	{};

\node (user3text)
	at (-1.5, -0.1)
	[]
	{User 3};
	
\draw
%	[decorate, decoration={bracket, mirror}, thick]
	[]
	(-5.9, 1.4)
	-- 
	++(-0.1, 0.0)
	-- node [rotate=90, above=.2em, align=center] {\( \jobrequestedset \)} 
	++(0.0, -1.1)
	--
	++(0.1, 0.0)
	;

\end{tikzpicture}

%\node (node1)
%	%at (+0.0*\xshift, +0.0*\yshift)
%	at (0, 0)
%	[draw]
%	{text};

%\draw[A0]
%	(node1) to (+1.0, +1.0);

%\draw[A1]
%	(node1)
%		edge []
%		node [below, align=center] {}
%	(-1.0, -1.0);

% -
	\caption{%
		The discrete {\scshape RA} problem considered in this paper.
		A scheduler assesses a set~\( \jobrequestedset \) of jobs that are assigned to a specific user and possess different properties.
		The scheduler must find a way to distribute a limited number~\( \totalresourceblocks \) of discrete resources among the users in order to optimize a utility function~\( \reward \).
		This process is repeated in each discrete \timesteptxt.
	}
	\label{fig:systemmodel}
\end{figure}

Several of the currently used medium access control protocols, chiefly among them the widely used \ac{ofdm}, separate the available resource bandwidth into discrete blocks.
Therefore, we consider a \ac{ra} problem as depicted in \reffig{fig:systemmodel}, where in every discrete \timesteptxt a scheduler at a base station is presented with a set of jobs \( \jobrequestedset \).
Each \jobtxt is destined for a specific user \( \userindex \in \numbersnatural \) from a total of \( \totalusers \in \numbersnatural \) users and comes with a number of properties, to be specified subsequently.
Based on the state of these properties, the scheduler must then distribute the total number of \( \totalresourceblocks \in \numbersnatural \) available discrete resource slots among the \( \totalusers \) users in order to maximize a utility function, to be defined in subsection \refsec{sec:problemstatement}.
The decision takes the form of an allocation vector
\begin{align}
	\label{eq:action}
	\actionvec = \left[ \percentageuser[1],\, \percentageuser[2],\, \dots,\, \percentageuser[\userindex],\, \dots,\, \percentageuser[\totalusers] \right]
	\text{ with } \sum_{\userindex = \num{1}}^{\totalusers} \percentageuser = 1.
\end{align}
According to this allocation, the system will then slot a fraction of user requests~\( \resourceblocksrequesteduser \in \numbersnatural\) into the available resources, starting from the oldest request per \usertxt.

%The set of all jobs cleanly divides into the user subsets \( { \jobrequesteduserset \subseteq \jobrequestedset } \) of jobs assigned to users~\( \userindex \) in \timesteptxt.
Jobs~\( \jobindex \) for each \usertxt are generated at the start of each \timesteptxt at a probability~\( \probnewjob \) per \usertxt.
Apart from the designated \usertxt, these jobs differ in three properties:
\begin{enumerate}
	\item Request size~\( \resourceblocksrequestedjob \in \numbersnatural \) in discrete resource blocks.
	The request size is initialized with a value \( { \resourceblocksrequestedjob \gets \resourceblocksrequesteduserinitial \sim \distributionuniform\left[1, \resourceblocksizesmax\right] } \) drawn from a uniform distribution.
	Upon allocation, the request size is decreased accordingly, and a count \( \resourcesscheduleduser \in \numbersnatural \) of all resource blocks scheduled to \usertxt in \timesteptxt is kept for performance metric calculation.
	Once the request size reaches zero, the job is removed from the set of requesting jobs~\( \jobrequestedset[\timeindex+\num{1}] \) for the next \timesteptxt;
	
	\item Delay~\( \jobdelay \in \numbersnatural \) in discrete time steps.
	The delay is initialized with \( { \jobdelay \gets \num{0} } \) and incremented at the end of each \timesteptxt where the request size has not reached zero.
	Should the delay then assume a value~\( { \jobdelay > \jobdelaymax } \), that \jobtxt is considered timed out, removed from the set~\( \jobrequestedset[\timeindex+\num{1}] \) of jobs requested for \timesteptxt, and instead added to a set~\( \jobstimedoutset \) of jobs \( \jobindex \) that have timed out in \timesteptxt;
	
	\item Priority status.
	Each jobs priority status is initialized as \emph{normal} upon generation.
	At the beginning of each \timesteptxt, at a low probability~\( \probpriorityjob \), one \jobtxt from the set~\( \jobrequestedset \) of all requesting jobs is assigned \emph{priority} status.
	If this \jobtxt is not scheduled by the end of that same \timesteptxt, it is considered as timed out regardless of its delay~\( \jobdelay \) and added to a set~\( \jobstimedoutpriorityset \) of priority jobs that have timed out in \timesteptxt.
	By definition, this set can have at most one member.
	We emphasize that these priority jobs constitute the significant rare events considered in this paper.
\end{enumerate}

The final piece of information for a scheduler to consider is the current channel state between the scheduler's base station and each \usertxt.
We model the connection by a Rayleigh fading channel, where the current channel power gain~\( { \left| \channelstateuser \right| \sim \text{Rayleigh}(\variancechannel) } \) to \usertxt in \timesteptxt is drawn from a Rayleigh distribution with variance~\( \variancechannel \).
In order to minimize the complexity of the simulation, we assume perfect knowledge of the channel state at the scheduler and opt to keep the channel power gain~\( \left|\channelstateuser\right| \) constant over all discrete resource blocks within a \timesteptxt.

\subsection{Problem Statement}
\label{sec:problemstatement}

We define three key metrics~\( \rewardsumrate,\, \rewardtimeout,\, \rewardtimeoutprio \) for the scheduler's consideration:
\begin{enumerate}
	\item \( \rewardsumrate \) is the Sum Rate over all users~\( \userindex \) under assumption of a Gaussian code book, \ie
	\begin{align}
		\rewardsumrate =
			\sum_{\userindex = \num{1}}^{\totalusers}
				\resourcesscheduleduser \cdot
				\log \left( 1 + \left| \channelstateuser \right|^{\num{2}} \text{SNR} \right)
	\end{align}
	with Signal-to-Noise ratio SNR, which we assume as constant over all users within this paper, and the count~\( \resourcesscheduleduser \) of all discrete resource blocks allocated to \usertxt in \timesteptxt;
	
	\item \( \rewardtimeout \) is the total number of timed out jobs~\( \jobindex \in \jobstimedoutset \) within \timesteptxt, \ie
	\begin{align}
		\rewardtimeout =
			\left| \jobstimedoutset \right|;
	\end{align}

	\item \( \rewardtimeoutprio \) is the total number of timed out \emph{priority} jobs~\( \jobindex \in \jobstimedoutpriorityset \) within \timesteptxt, \ie
	\begin{align}
		\rewardtimeoutprio =
			\left| \jobstimedoutpriorityset \right|.
	\end{align}
\end{enumerate}

As we wish to simultaneously maximize the Sum Rate~\( \rewardsumrate \) and minimize the time outs~\( \rewardtimeout \) and \( \rewardtimeoutprio \), we define the optimization goal~\( \reward \) to be a weighted sum of the three key metrics,
\begin{align}
	\label{eq:reward}
	\reward =
		\weightsumrate \rewardsumrate
		+ \weighttimeout \rewardtimeout
		+ \weighttimeoutprio \rewardtimeoutprio
	,
\end{align}
where the weights~\(\weightsumrate,\, \weighttimeout,\, \weighttimeoutprio\) scale the relative importance of the metrics~\( \rewardsumrate,\, \rewardtimeout,\, \rewardtimeoutprio \).
The following section will detail the design of a \ac{rl} scheduler that learns to approximately optimize the weighted sum metric~\( \reward \) by trial-and-error.
%

%%%%%%%%%%%%%%%%%%%%%%%%%%%%%%%%%%%%%%%%%%%%%%%%%%%%%%%%%%
% REMEMBER STORYTELLING
%%%%%%%%%%%%%%%%%%%%%%%%%%%%%%%%%%%%%%%%%%%%%%%%%%%%%%%%%%

\section{Learned Schedulers}
\label{sec:learnedschedulers}

In order to learn effective scheduling in the presence of Black Swan Events, we implement a vanilla deep actor critic learner and extend it with two mechanisms from the domain of continual learning.
This section will first describe the components of the learner, how they interact, as well as the training algorithm that iteratively optimizes performance.
In the two following subsections, we present \ac{ewc}~\cite{kirkpatrick_overcoming_2017} and \ac{gem}~\cite{lopez-paz_gradient_2017}, and demonstrate how we integrate them with the learner to fortify against Black Swan Events.

\subsection{Actor-Critic Scheduler Design \& Training}
\label{sec:actorcritic}

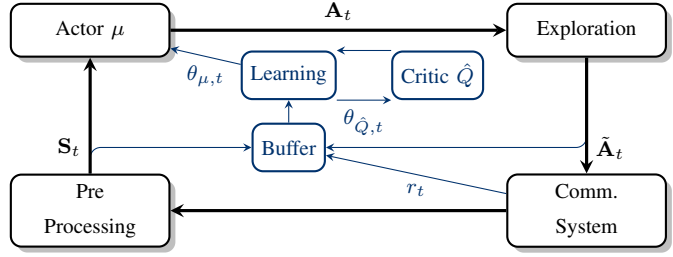
\begin{figure}[!t]
	\centering
	\begin{tikzpicture}[scale=1.0, font={\footnotesize}]

\tikzstyle{A0} = [-, >={stealth}, rounded corners]
\tikzstyle{A1} = [->, >={stealth}, rounded corners]
\tikzstyle{A2} = [<->, >={stealth}, rounded corners]

\tikzstyle{A0training} = [A0, color=uniblue1]
\tikzstyle{A1training} = [A1, color=uniblue1]
\tikzstyle{A2training} = [A2, color=uniblue1]

\tikzstyle{A0evaluation} = [A0, thick]
\tikzstyle{A1evaluation} = [A1, very thick]
\tikzstyle{A2evaluation} = [A2, thick]

\tikzstyle{box} = [draw, rounded corners, align=center]
\tikzstyle{boxtraining} = [box, thick, color=uniblue1, minimum height=1.8em, fill=white,]
\tikzstyle{boxevaluation} = [box, thick, minimum width=6em, minimum height=2em, fill=white, drop shadow]

\newcommand{\xshift}{6.6}
\newcommand{\yshift}{2.4}

\node (actor)
	at (0.0, 0.0)
	[boxevaluation]
	{Actor \( \actornetworkvec \)};
	
\node (exploration)
	at (\xshift, 0.0)
	[boxevaluation]
	{Exploration};
	
\node (environment)
	at (\xshift, -1.0*\yshift)
	[boxevaluation]
	{Comm.\\System};
	
\node (preproc)
	at (0.0, -1.0*\yshift)
	[boxevaluation]
	{Pre\\Processing};
	
\node (buffer)
	at (.4*\xshift, -.65*\yshift)
	[boxtraining]
	{Buffer};
	
\node (critic)
	at (.7*\xshift, -.25*\yshift)
	[boxtraining]
	{Critic \( \criticnetworksca \)};
	
\node (learning module)
	at (.4*\xshift, -.25*\yshift)
	[boxtraining]
	{Learning};

\draw[A1training]
	(exploration.south) |-
	node [right] {\color{black}\footnotesize \( \actionvecnoisy \)}
	(buffer.east);
	
\draw [A1training]
	(environment) --
	node [below] {\footnotesize \( \reward \)}
	(buffer);
	
\draw [A1training]
	(preproc.north) |-
	node [left] {\color{black}\footnotesize \( \statevec \)}
	(buffer.west);

\draw [A1training]
	(learning module.south east) --
	node [align=center, below] {\footnotesize \( \parameterscritic \)}
	(critic.south west);

\draw [A1training]
	(critic.north west) --
	(learning module.north east);
	
\draw [A1training]
	(learning module) --
	node [align=center, below] {\footnotesize \( \parametersactor \)}
	(actor);

\draw [A1training]
	(buffer) --
	(learning module);
	
\draw [A1evaluation]
	(actor) --
	node [above] {\footnotesize \( \actionvec \)}
	(exploration);

\draw [A1evaluation]
	(exploration) to (environment);
	
\draw [A1evaluation]
	(environment) to (preproc);
	
\draw [A1evaluation]
	(preproc) to (actor);

\end{tikzpicture}
	\caption{%
		The actor-critic learned scheduler's process flow contains two loops.
		The outer, independent loop, marked in boldface black, infers scheduling decisions~\( \actionvec \) based on the current system state~\( \statevec \) and, during the training phase, generates data samples for the learning process.
		The exploration module transforms the scheduling decisions by adding noise so as to generate a more rich training data set.
		In parallel, the inner loop, marked in blue, collects the data samples generated by the outer loop in a memory buffer, uses them to update a critic \ac{nn} to better estimate the goodness of a scheduling decision and, based on that, also updates the actor \ac{nn} to output better scheduling decisions.
	}
	\label{fig:processflow}
\end{figure}

The actor-critic learner used in this paper consists of a pre-processor, two \ac{nn} (\emph{actor}~\( \actornetworkvec \) with parameters~\( \parametersactor \) and \emph{critic}~\( \criticnetworksca \) with parameters~\( \parameterscritic \)), an exploration module, a memory buffer, and a learning module.
All components contribute to the learning process, whereas for inference only pre-processor and actor \ac{nn}~\( \actornetworkvec \) are involved.
While the actor \ac{nn}~\( \actornetworkvec \) infers the actual scheduling decision~\( \actionvec \) (see \refeq{eq:action}), the role of the critic \ac{nn}~\( \criticnetworksca \) is to estimate the goodness of an action~\( \actionvec \) in a given situation, or in other words, approximate the hidden system dynamics.
\reffig{fig:processflow} illustrates the process flow of these components which we will describe in more detail in the following.

In every \timesteptxt, the pre-processor first summarizes all information pertinent to scheduling decisions into a system state vector~\( \statevec \) that can be used as an input to the \ac{nn}.
Specifically, the state vector contains four features per user~\( \userindex \):
\begin{enumerate}
	\item Resources requested by \usertxt in \timesteptxt, normalized by total available resources~\( \totalresourceblocks \),
	\begin{align}
		\textstyle
		\statescauser{1} =
			\frac{1}{\totalresourceblocks}
			\sum_{\jobindex\in\jobrequesteduserset} \resourceblocksrequestedjob
		;
	\end{align}

	\item Priority resources requested by \usertxt in \timesteptxt, normalized by total available resources~\( \totalresourceblocks \),
	\begin{align}
		\textstyle
		\statescauser{2} =
			\frac{1}{\totalresourceblocks}
			\sum_{\jobindex\in\{\jobrequesteduserset,\, \jobindex \text{ is prio}\}} \resourceblocksrequestedjob
		;
	\end{align}

	\item Instantaneous channel power gain of \usertxt at \timesteptxt,
	\begin{align}
		\textstyle
		\statescauser{3} = 
			\left| \channelstateuser \right|^{\num{2}}
		;
	\end{align}

	\item Maximum delay of \usertxt at \timesteptxt, normalized by the maximum allowed delay~\( \jobdelaymax \),
	\begin{align}
		\textstyle
		\statescauser{4} = 
			\max_{\jobindex\in\jobrequesteduserset} \jobdelay / \jobdelaymax
		.
	\end{align}
\end{enumerate}

This state vector~\( \statevec \) of length~\( \num{4}\totalusers \) is then used as input to the actor \ac{nn}~\( \actornetworkvec \), which, based on its current parametrization~\( \parametersactor \), will output an allocation solution~\( \actionvec \) in the form of \refeq{eq:action} that contains the relative share of the total resources~\( \totalresourceblocks \) that each \usertxt is allotted.
In order to generate a rich data set to learn from, that action vector~\( \actionvec \) is further modified by the exploration module.
The module first generates a random vector \( \explorationvec \) of the same length as \( \actionvec \) with its entries~\( { \explorationvecsca \sim \distributionuniform[\num{0}, \num{1}] } \) drawn from a random uniform distribution.
This vector is then mixed with the action vector~\( \actionvec \) according to a momentum parameter~\( \explorationmomentum \),
\begin{align}
	\actionvecnoisy =
		\explorationmomentum \explorationvec
		+ (\num{1} - \explorationmomentum) \actionvec
	.
\end{align}
As we move away from training towards inference, \( \explorationmomentum \) is scaled down progressively so as to perturb the learned scheduler's inference less and less.
The resulting noisy action vector~\( \actionvecnoisy \) is re-normalized and forwarded to the communication system, where it progresses the system state as described in \refsec{sec:systemmodel}, resulting in a success metric~\( \reward \) and a new system state~\( \statevec[\timeindex+\num{1}] \).
The data tuple of state~\( \statevec \), action~\( \actionvec \) and results~\( \reward \) is saved in the experience buffer.

The learning module may now leverage this collected data to update the actor \ac{nn}~\( \actornetworkvec \) to output better allocation decisions that generate a higher expected reward metric~\( \reward \).
In order to do this, first, the critic \ac{nn}~\( \criticnetworksca \) is updated.
The critic \ac{nn} takes as input a combination of state~\( \statevec \) and action~\( \actionvec \) and estimates the expected reward metric, \( { \criticnetworksca_{\parameterscritic}(\statevec, \actionvec) = \rewardestimate } \).
Therefore, we wish to minimize the distance between reward metric estimate and the actual reward metrics recorded in the experience buffer,
\begin{align}
	\label{eq:losscritic}
	\losscritic =
		\left( \criticnetworksca_{\parameterscritic}(\statevec, \actionvec) - \reward \right)^{\num{2}}
	.
\end{align}
The critic \ac{nn}'s estimate~\( \rewardestimate \) of the reward metric~\( \reward \) can then be used as a target for updating the actor \ac{nn}~\( \actornetworkvec \).
Given a good enough approximation, the action~\( { \actionvec = \actornetworkvec(\statevec) } \) that maximizes the expected reward metric~\( { \criticnetworksca_{\parameterscritic}(\statevec, \actionvec) = \rewardestimate } \) also approximately maximizes the estimated reward~\( \reward \).
Therefore, our optimization target for the actor \ac{nn} will be
\begin{align}
	\label{eq:lossactor}
	\lossactor =
		- \criticnetworksca(\statevec, \actornetworkvec_{\parametersactor}(\statevec))
\end{align}
for a given state~\( \statevec \) from the memory buffer.
In practice, the parameters~\( \parameterscritic \), \( \parametersactor \) of critic and actor \ac{nn} respectively are updated to minimize both loss functions \refeq{eq:losscritic} and \refeq{eq:lossactor} on batches of experiences from the memory buffer with \ac{sgd}-like optimization.
While \ac{sgd} has proven to be very efficient, it brings along problems when facing a situation where significant state-action pairs are underrepresented in the learning data set.
Further, \ac{sgd} does not have any built-in mechanisms to prevent forgetting, \ie if a particular state-action combination is phased out of the learning data set, the information contained may eventually be overwritten by subsequent training steps.
Consequently, in the following two subsections, we will present and incorporate two methods from continual learning that harden the actor-critic scheduler against these issues.

\subsection{\acf{ewc}}
\label{sec:ewc}

Both multi-task learning methods presented in this paper decompose the training process into two sequential stages: 1)~First learning to solve the problem of Black Swan Events exclusively to satisfaction; 2)~Then learning to optimize overall system performance~\( \reward \) while constrained to try to preserve performance on the first stage.
The first of these methods, which we first presented in~\cite{gracla_robust_2022}, makes use of \acl{ewc} as shown in~\cite{kirkpatrick_overcoming_2017}. 
The main idea is to add an elastic penalty term to the actor \ac{nn} optimization objective, \ie the loss function~\refeq{eq:lossactor}, that constrains the learning of overall performance optimization in step~2) to solutions that are ``nearby'' a solution that is known to adequately handle significant rare events, \ie the solution found in step~1).
As the \ac{nn} \emph{parameters} encode the scheduler's behavior, the authors extract two indicators for ``nearbiness'' from the actor \ac{nn}~\( \actornetworkvec \) after the first stage: 1)~Each parameter~\( \parameterindex \)'s final values, denoted as~\( \parameteractoranchor \) from here on; 2)~Each parameter~\( \parameterindex \)'s Fisher information~\( \fisherinformationparameter \).
The Fisher information~\( \fisherinformationparameter \) describes the local flatness of the optimization landscape surrounding a given parameter~\( \parameterindex \); it therefore describes the sensitivity of the \ac{nn} behavior to changes in that specific parameter~\( \parameterindex \).
Adjusting a parameter surrounded by a flat area on the optimization landscape is less likely to significantly change the pre-learned behavior, therefore, this parameter is more attractive and safe to adjust than a parameter on a steep slope.

Accordingly, for the second learning stage of optimizing overall system performance, a penalty term is added to the actor \ac{nn} optimization objective \refeq{eq:lossactor}.
For all \( \totalparameters \)~parameters, this penalty considers the Euclidian distance between the parameters recorded from the first task~\( \parameteractoranchor \) and the current parameters~\( \parameteractor \), weighted by the Fisher information~\( \fisherinformationparameter \).
It is constructed as
\begin{align}
	\label{eq:lossewc}
	\lossewc = 
		\weightanchor \sum_{\parameterindex=\num{1}}^{\totalparameters}
			\fisherinformationparameter \left( \parameteractor - \parameteractoranchor \right)^{\num{2}}
\end{align}
with the weighting factor~\( \weightanchor \) added to scale the \ac{ewc} penalty's importance.
This penalty term is minimized by keeping current parameters~\( \parameteractor \) close in value to the parameters~\( \parameteractoranchor \) that are known to encode proper handling of priority events, with a focus on those parameters with high Fisher information~\( \fisherinformationparameter \), \ie those parameters particularly sensitive to change.
It therefore pulls the learning process towards solutions that cause less perturbation to the previously learned solution.
Approximations for the Fisher information~\( \fisherinformationparameter \) can often be acquired at no additional computational cost, as modern \ac{sgd} implementations like the widely used Adam optimizer~\cite{kingma_adam_2015} already make use of it internally.
For a more in-depth look at the goodness of Fisher information approximations, we refer to~\cite{achille_critical_2019}.

\subsection{\acf{gem}}
\label{sec:gem}

Like the \ac{ewc} method in the previous chapter, \ac{gem}~\cite{lopez-paz_gradient_2017} also splits the learning process into the two tasks of first learning how to deal with the rare priority events well, and secondly optimizing the overall system performance~\( \reward \) with an additional constraint to limit performance degradation on the first task.
\ac{gem} constructs this constraint based on a sample of data retained from the first training stage, \ie in our case, a set of \( \gemtotalnumsamples \)~data samples that contain information on how to specifically deal with the problematic events.
Given this data set, we may calculate the gradients for the actor and critic \ac{nn} according to optimization objectives~\refeq{eq:losscritic}, \refeq{eq:lossactor}.
The authors in~\cite{lopez-paz_gradient_2017} now argue that, assuming the data set is representative and the optimization landscapes are locally linear, we are able to check the alignment of the gradient vectors~\( { \nabla_{\mathbf{\internalparameter}}\mathcal{\loss}_{\text{prio}} = \gemgradientold \in \numbersreal^{\num{1}\times\totalparameters} } \) given the retained data set and the gradient vectors~\( {\nabla_{\mathbf{\internalparameter}}\mathcal{\loss}_{\text{curr}} =  \gemgradientcurrent \in \numbersreal^{\num{1}\times\totalparameters} } \) given the current training data from the regular learning process as described in \refsec{sec:actorcritic} via the scalar product
\begin{align}
	\label{eq:gemcondition}
	\left\langle \gemgradientold,\, \gemgradientcurrent \right\rangle
	.
\end{align}
Local linearity may be assumed when using iterative, \ac{sgd}-like optimizers with small step sizes.
If the above scalar product is greater than zero, both gradient vectors are aligned and the parameter update using the gradient on the current training data, \( \gemgradientcurrent \), is unlikely to negatively affect performance on the priority scheduling task.
On the other hand, if they are not aligned, the authors suggest to project the gradient vector~\( \gemgradientcurrent \) geometrically to the least perturbed alternative gradient vector~\( \gemgradientnew \) that has a positive scalar product with the priority task gradient~\( \gemgradientold \), or mathematically,
\begin{align}
	\min_{\gemgradientnew}
		& \frac{\num{1}}{\num{2}} \left\| \gemgradientnew - \gemgradientcurrent \right\|^{\num{2}}_{\num{2}} \nonumber\\
	\text{s.t. }
		&\left\langle \gemgradientold,\, \gemgradientnew \right\rangle \geq \num{0}
	.
\end{align}
The authors then show that this problem can be formulated as a quadratic program, to be solved by a numerical solver.
Specifically, they show that this problem can be expressed as an optimization over an amount of variables equal to the amount of constraint data sets, which in our case is only one.
The problem to be solved numerically is
\begin{align}
	\nonumber
	\label{eq:gemnumerical}
	\gemoptimizationvariableoptimal =
		\min_{\gemoptimizationvariable}
			& \frac{\num{1}}{\num{2}}
			\gemoptimizationvariable^{\num{2}} \left\| \gemgradientold \right\|_{\num{2}}^{\num{2}}
			+ \gemgradientcurrent \gemgradientold^{\text{T}} \gemoptimizationvariable \\
			\text{s.t. } & \gemoptimizationvariable \geq \num{0}
\end{align}
From the result~\( \gemoptimizationvariableoptimal \) of this numerical optimization, the projected gradient vector~\( \gemgradientnew \) can be constructed as a linear combination of the gradient from the retained data set~\( \gemgradientold \) and from the current data set~\( \gemgradientcurrent \) as
\begin{align}
	\gemgradientnew =
		\gemgradientold \gemoptimizationvariableoptimal + \gemgradientcurrent
	.
\end{align}
%As the assumptions of linear locality and representativeness of the prior data set may not hold all of the time in reality, the authors suggest to add a small tuning variable~\( \gemweight \) onto ~\( \gemoptimizationvariableoptimal \).
By this mechanism, \ac{gem} tries to explicitly encourage learning transfer between the two stages in both directions: backward transfer, \ie updates on the current data set improve performance on handling priority events, and forward transfer, \ie inclusion of the priority data improving performance on the overall optimization.
%

%%%%%%%%%%%%%%%%%%%%%%%%%%%%%%%%%%%%%%%%%%%%%%%%%%%%%%%%%%
% REMEMBER STORYTELLING
%%%%%%%%%%%%%%%%%%%%%%%%%%%%%%%%%%%%%%%%%%%%%%%%%%%%%%%%%%

\section{Evaluation}
\label{sec:experiments}

In this section, we compare the performance of the learned schedulers on discrete \ac{ra} as described in \refsec{sec:systemmodel}.
All schedulers will be evaluated in overall performance, \ie maximizing the average weighted performance sum~\( \reward \) from~\refeq{eq:reward}, as well as their handling of priority events.
The default simulation that all schedulers are evaluated on has a low probability~\( \probpriorityjob = \num{1} / \num{10000} \) of encountering a priority event per \timesteptxt.

\subsection{Implementation Details}
\label{sec:implementationdetails}

We implement the simulation and training loop in python on generic hardware, using the TensorFlow library and the Adam optimizer~\cite{kingma_adam_2015} with a learning rate or step size~\( \learningrate \).
Numerical solving of~\refeq{eq:gemnumerical} is done through the CVXOPT library.
\reftab{tab:parameters} lists the most important simulation parameters, the full code implementation is available at~\cite{gracla_code_2022}.
%For \ac{gem}, we did not find the tuning factor~\( \gemweight \) to be beneficial and therefore set it to \( {\gemweight = \num{0}} \).

\begin{table}[!t]
	\renewcommand{\arraystretch}{1.3}
	\caption{Selected Simulation Parameters}
	\label{tab:parameters}
	\centering
	\rowcolors{2}{white}{uniblue1!5} 
	\begin{tabular}{llll}
		\hline
		Total Users \( \totalusers \) & \num{5}
			&
			Total Resources \( \totalresourceblocks \) & \num{10}
		\\
		Job Creation Prob. \( \probnewjob \) & \num{0.5}
			&
			Prio. Event Prob. \( \probpriorityjob \) & \( \num{10}^{\num{-4}} \)
		\\
		SNR & \SI{10}{dB}
			&
			Rayleigh Scale \( \variancechannel \) & \num{0.3}
		\\
		Job Max Size \( \resourceblocksrequesteduserinitial \) & \num{7}
			&
			Max Delay \( \jobdelaymax \) & \num{5} 
		\\
		Training Episodes \( \totalepisodes \) & 30
			&
			Steps per Episode \( \totaltimesteps \) & \num{10000}
		\\
		Weight Sum Rate \( \weightsumrate \) & \num{1}
			&
			Episodes \( \explorationmomentum \rightarrow \num{0} \) & \SI{50}{\percent}
		\\
		Weight Time Out \( \weighttimeout \) & \num{-1}
			&
			Learning Rate \( \learningrate \) & \( 10^{-4} \)
		\\
		Weight Prio. \( \weighttimeoutprio \) & \num{-5}
			&
			\ac{nn} Hidden Nodes & \( \num{3} \times \num{128} \)
		\\
		\hline
	\end{tabular}
\end{table}

The basic training and evaluation loop for all schedulers is as follows.
A \ac{nn} is trained for \( \totaltimesteps \) time steps, after which the simulation is reset, and the process is repeated for \( \totalepisodes \) training episodes.
In each time step, an inference is made, the experience saved, and one training step is performed according to \refsec{sec:learnedschedulers}, including adaptations made for \ac{gem} and \ac{ewc} where applicable.
After the full amount of training steps, the \ac{nn}'s final configuration is frozen and used for evaluation on the baseline simulation, for a total of \( \totalepisodes = \num{5} \) episodes with a large number of \( { \totaltimesteps = \num{200000} } \) time steps~\( \timeindex \) each, giving ample opportunity to observe behavior on priority events even at their rare occurrence.
All training and evaluations are repeated three times, with their results averaged, to ascertain that the performance is stable and reliable.

\subsection{Results}
\label{sec:results}

\begin{figure}[!t]
	\centering
	\includegraphics{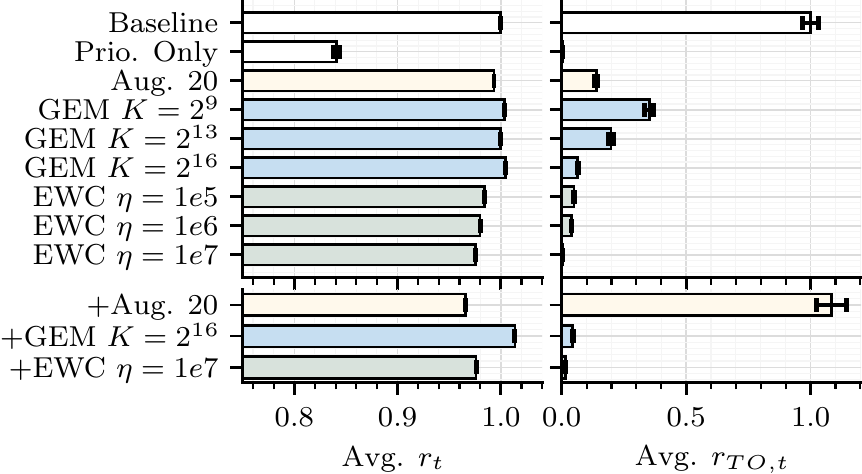}
	\caption{%
		Mean results when using the learned schedulers in evaluation, normalized to the baseline.
		The left graph shows the performance in overall optimization goal~\( \reward \), where a higher score is better, while the right graph focuses on priority event handling exclusively, where lower is better.
		Methods are grouped by color, blue denotes the use of {\scshape GEM}, green for {\scshape EWC}, and yellow for schedulers that were not extended but trained on an augmented simulation, all other schedulers are white.
		Black bars represent the variance.
		Note that the left graph's axis is scaled to highlight the relevant area.
	}
	\label{fig:results}
\end{figure}

We train five types of schedulers:
\begin{enumerate}
	\item The \emph{Baseline} scheduler is using a randomly initialized \ac{nn} and is trained directly on the same simulation configuration that all schedulers are evaluated on, \ie \( { \probpriorityjob = \num{1} / \num{10000} } \);
	\label{item:baseline}

	\item The \emph{Prio. Only} scheduler is randomly initialized and is trained on a simulation with exclusively priority events, \( {\probpriorityjob = \num{1}} \);
	\label{item:critical}

	\item \emph{\small GEM} schedulers are initialized with the final state of the Prio. Only scheduler, then trained on the baseline simulation with \( {\probpriorityjob = \num{1} / \num{10000}} \).
	We evaluate three choices of sample memory size~\( \gemtotalnumsamples \in [ \num{2}^{\num{9}}, \num{2}^{\num{13}}, \num{2}^{\num{16}} ] \);
	\label{item:gem}

	\item \emph{\small EWC} schedulers similarly are initialized with the final state of the Prio. Only scheduler, then trained on the baseline simulation with \( {\probpriorityjob = \num{1} / \num{10000}} \).
	We evaluate three choices of anchoring weight~\( \weightanchor \in [1\text{e}5, 1\text{e}6, 1\text{e}7 ] \);
	\label{item:ewc}

	\item For a benchmark, we train a randomly initialized \ac{nn} on an \emph{augmented} simulation, \ie the probability to encounter priority events is artificially increased by a significant degree, to \( {\probpriorityjob = \num{0.2}} \).
	For a fair comparison, to compensate for the pre-training of \ac{ewc}, \ac{gem}, these schedulers are trained for twice the amount of training episodes.
	\label{item:aug}
\end{enumerate}
The results achieved in evaluation are illustrated in the upper section of \reffig{fig:results}.
From schedulers \refitem{item:baseline}, \refitem{item:critical} we can see the effects of naive training: Focusing only on the overall performance \( \reward \) does not properly learn the handling priority events, while exclusive focus on priority events during training leads to a drop in overall performance in evaluation.
The goal of all other schedulers will be to fill this performance gap while keeping priority event time outs as low as possible.
The benchmark \refitem{item:aug}, training on an augmented simulation with significantly increased amount of priority samples, shows that striking this balance is indeed possible.
Regarding \ac{ewc}, \ac{gem}, we report that they, too, can leverage their respective multi-task learning mechanisms to maintain proper handling of priority events while closing the overall performance gap.
We clearly highlight the impact of the relative tuning parameters, the weight \( \weightanchor \) for \ac{ewc} as shown in \refeq{eq:lossewc}, and the memory size~\( \gemtotalnumsamples \) for \ac{gem}, both increasing in effect with increasing magnitude.
\ac{ewc} suffers from a relatively larger drop in overall performance, but has an easier time with priority events over all configurations, while \ac{gem} maintains high overall performance, but requires a large amount of memory samples.
We attribute this to the non-continuous nature of mapping discrete resources with a continuous output, which may violate the assumption of local linearity and necessitates high amounts of samples for a representative data set.
While a large memory size~\( \gemtotalnumsamples \) comes with a performance cost in training, as all \( \gemtotalnumsamples \)~samples have to be evaluated in each training step, we do not consider this restriction to be overly punishing, as the training happens off-line and can be carried out on dedicated hardware.

While the performance of \ac{ewc}, \ac{gem} may not look overly impressive compared to the augmented simulation, recall that these methods bring further benefits.
To emphasize this, we select the Aug. 20 scheduler as well as one choice of \ac{ewc}, \ac{gem} each, unfreeze the \acp{nn}, and continue their training on a simulation with \( {\probpriorityjob = \num{0}} \), encountering no further priority events.
The results after evaluation are displayed in the lower section of \reffig{fig:results}, marked with a plus sign.
We clearly see the effect of catastrophic forgetting on the Aug. 20 scheduler, which entirely unlearns how to handle priority events.
Meanwhile, the \ac{ewc} scheduler maintains performance on both metrics, while the \ac{gem} mechanism can even leverage positive transfer learning, achieving the best overall performance out of all schedulers.
Recall further that, as a simulation model comes closer to reality, augmenting a simulation to the desired effect tends to become more and more complex and the augmented simulations data distribution moves further from reality, while \ac{ewc}, \ac{gem} remain at a single control parameter and are trained directly on the true simulation model.
%

%%%%%%%%%%%%%%%%%%%%%%%%%%%%%%%%%%%%%%%%%%%%%%%%%%%%%%%%%%
% REMEMBER STORYTELLING
%%%%%%%%%%%%%%%%%%%%%%%%%%%%%%%%%%%%%%%%%%%%%%%%%%%%%%%%%%

\section{Conclusion}
\label{sec:conclusions}

This paper proposes to decompose a discrete \acl{ra} challenge with Black Swan Events into a two-stage \acl{rl} problem, thereby enabling the use of two methods from multi-task learning, \acf{ewc} and \acf{gem}.
We show that this approach is well able to handle the priority events while offering advantages in robustness over the state-of-the-art approach of changing the training data distribution, at the cost of increased training complexity.
%

% \acresetall to reset the "first use" flag of acronym package
% \acused{name} to set "first use" flag, good for well-known acronyms

% \ac{name}  default command: full name with short in brackets on first use only
% \acf{name}  give full name with acronym in brackets
% \acs{name}  give short name only
% \acl{name}  long name only
% commands for plural also exist

\begin{acronym}
	[\textsc{\small OFDMA}]  % put the longest acronym into this option box
	
	\acro{sota}[\textsc{\small SotA}]{State of the Art}
	\acro{dof}[\textsc{\small DoF}]{Degree of Freedom}
	
	\acro{3gpp}[\textsc{\small 3GPP}]{3rd Generation Partnership Project}
	
	\acro{ra}[\textsc{\small RA}]{Resource Allocation}
	
	\acro{urllc}[\textsc{\small URLLC}]{Ultra Reliable and Low Latency Communications}
	\acro{embb}[\textsc{\small eMBB}]{enhanced Mobile Broadband}
	\acro{mmtc}[\textsc{\small mMTC}]{massive Machine Type Communication}
	
	\acro{tx}[\textsc{\small Tx}]{Transmit}
	\acro{rx}[\textsc{\small Rx}]{Receive}
	
	\acro{csit}[\textsc{\small CSIT}]{Channel State Information at Transmitter}
	\acro{ber}[\textsc{\small BER}]{Bit Error Rate}
	
	\acro{sgd}[\textsc{\small SGD}]{Stochastic Gradient Descent}
	\acro{mmse}[\textsc{\small MMSE}]{Minimum Mean Squared Error}
	
	\acro{dtft}[\textsc{\small DTFT}]{Discrete Time Fourier Transform}
	\acro{dft}[\textsc{\small DFT}]{Discrete Fourier Transform}
	
	\acro{zf}[\textsc{\small ZF}]{Zero Forcings}
	\acro{awgn}[\textsc{\small AWGN}]{Additive White Gaussian Noise}
	\acro{los}[\textsc{\small LOS}]{Line of Sight}
	
	\acro{ofdm}[\textsc{\small OFDM}]{Orthogonal Frequency Division Multiplex}
	\acro{ofdma}[\textsc{\small OFDMA}]{Orthogonal Frequency Division Multiplex}
	\acro{noma}[\textsc{\small NOMA}]{Non Orthogonal Multiple Access}
	
	\acro{siso}[\textsc{\small SISO}]{Single Input Single Output}
	\acro{simo}[\textsc{\small SIMO}]{Single Input Multiple Output}
	\acro{miso}[\textsc{\small MISO}]{Multiple Input Single Output}
	\acro{mimo}[\textsc{\small MIMO}]{Multiple Input Multiple Output}
	\acro{snr}[\textsc{\small SNR}]{Signal to Noise Ratio}
	
	\acro{ml}[\textsc{\small ML}]{Machine Learning}
	\acro{dl}[\textsc{\small DL}]{Deep Learning}
	\acro{rl}[\textsc{\small RL}]{Reinforcement Learning}
	\acro{nn}[\textsc{\small NN}]{Neural Network}
	
	\acro{dqn}[\textsc{\small DQN}]{Deep Q Network}
	\acro{ddpg}[\textsc{\small DDPG}]{Deep Deterministic Policy Gradient}
	\acro{ppo}[\textsc{\small PPO}]{Proximal Policy Optimization}
	
	\acro{ewc}[\textsc{\small EWC}]{Elastic Weight Consolidation}
	\acro{gem}[\textsc{\small GEM}]{Gradient Episodic Memory}
	
	\acro{ntn}[\textsc{\small NTN}]{Non Terrestrial Networks}
	\acro{geo}[\textsc{\small GEO}]{Geostationary Earth Orbit}
	\acro{leo}[\textsc{\small LEO}]{Low Earth Orbit}
	
\end{acronym}%
\bibliographystyle{ref/IEEEtran}%
{%
% 	% prevent widows, orphans
% 	\makeatletter  
% 		\clubpenalty=10000  
% 		\@clubpenalty=\clubpenalty
% 		\widowpenalty=10000
% 	\makeatother
% 	%
	\bibliography{ref/IEEEabrv,ref/references}%
}%

\end{document}